\documentclass{llncs}

\usepackage{graphicx}

\usepackage{url}
\usepackage{booktabs}
\usepackage{multirow}
\usepackage{siunitx}
\usepackage{hyperref}

\begin{document}
\title{SemiMemes: A Semi-supervised Learning Approach for Multimodal Memes Analysis}
%
%
\author{Pham Thai Hoang Tung\inst{1}\and
Nguyen Tan Viet\inst{1}\and Ngo Tien Anh\inst{1}\and
Phan Duy Hung\inst{1}}
%
%
\institute{FPT University, Hanoi, Vietnam \\
\email{\{tungpthhe141564,vietnthe153763,anhnthe141442\}@fpt.edu.vn} \email{hungpd2@fe.edu.vn}}
\maketitle              
\begin{abstract}
The prevalence of memes on social media has created the need to sentiment analyze their underlying meanings for censoring harmful content. Meme censoring systems by machine learning raise the need for a semi-supervised learning solution to take advantage of the large number of unlabeled memes available on the internet and make the annotation process less challenging. Moreover, the approach needs to utilize multimodal data as memes' meanings usually come from both images and texts. This research proposes a multimodal semi-supervised learning approach that outperforms other multimodal semi-supervised learning and supervised learning state-of-the-art models on two datasets, the Multimedia Automatic Misogyny Identification and Hateful Memes dataset. Building on the insights gained from Contrastive Language-Image Pre-training, which is an effective multimodal learning technique, this research introduces SemiMemes, a novel training method that combines auto-encoder and classification task to make use of the resourceful unlabeled data.

\end{abstract}

\section{Introduction}
An "internet meme" or "meme" is a concept, well-known phrase, pattern, or action that is distributed through the internet~\cite{borzsei2013makes}. The most prevalent type of internet meme is composed of a picture and a brief caption overlaid on top of it, which is what our research primarily concentrates on. In order to comprehend the significance of memes, it is occasionally necessary to grasp the significance of both the image and the text, and then connect them together.

As memes have become more popular, there is a trend of creating memes that are not meant to be funny or amusing, but to express irony or to spread negative content on sensitive topics such as discrimination, race, gender, religion, or politics. Consequently, social media platforms are paying attention to the issue of "memes sentiment analysis" to prevent the dissemination of memes that contain harmful content.

\begin{figure}
\includegraphics[width=\textwidth]{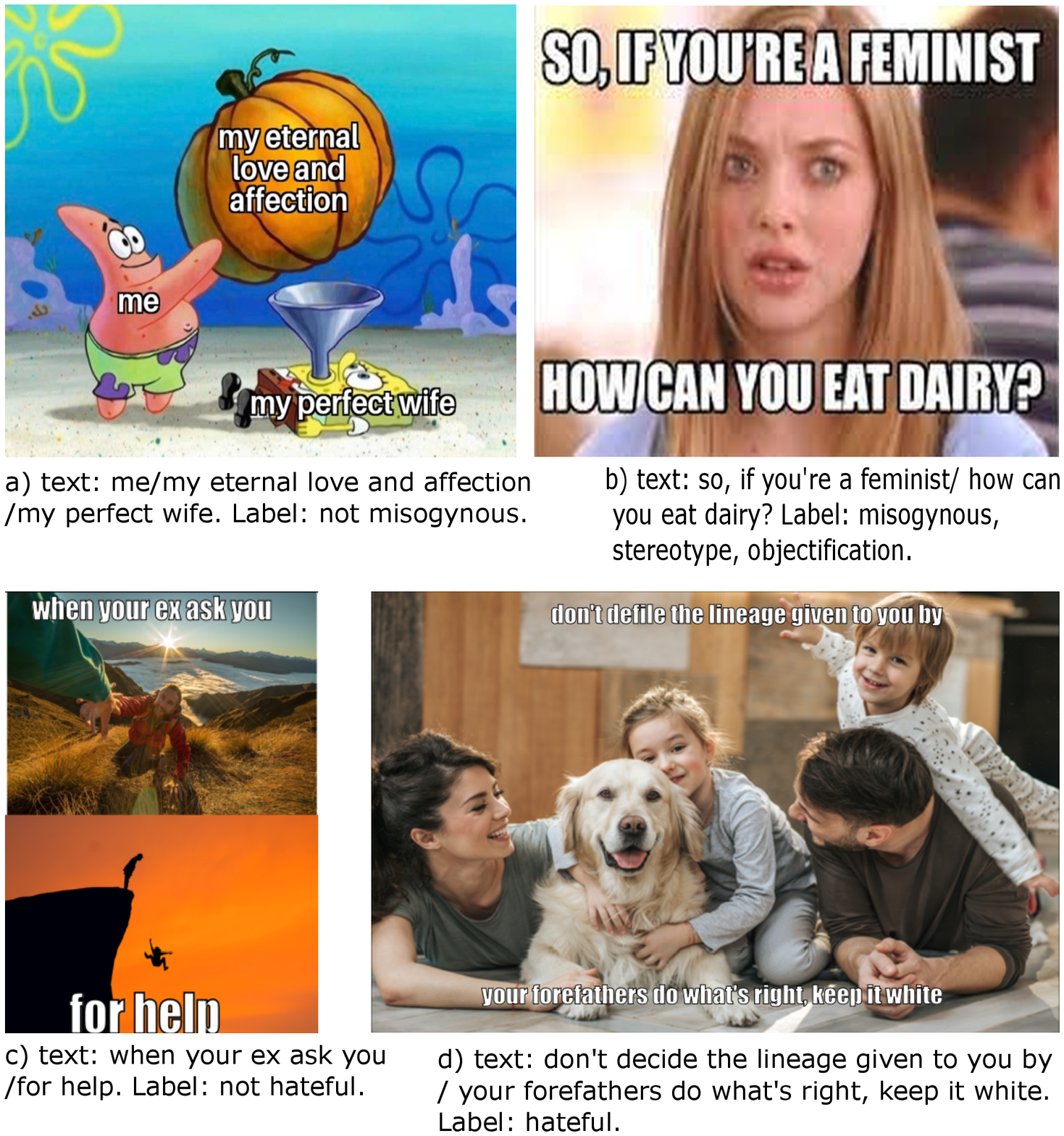}
\caption{
Illustrative examples from the Multi-
media Automatic Misogyny Identification dataset (a,b) and examples from the Hateful Memes dataset (c,d).
}
\label{figMeme}
\end{figure} 

In earlier times, harmful content was mainly conveyed through text that could be detected using Natural Language Processing. However, identifying harmful content hidden in memes has become a challenging task. The examples presented in Fig.~\ref{figMeme} demonstrate that the "memes sentiment analysis" approach to detect harmful content is complicated as it necessitates understanding and merging the visual and textual aspects of the meme. Even humans require time to comprehend it, so it is even more challenging to teach machines to recognize such content. Additionally, manually labeling data is exhaustive and could be subjective due to labeling participants, leading to disputes and arguments, as each person might interpret the meme's content differently~\cite{fersini-etal-2022-semeval,pmlr-v5-goldberg09a,pramanick-etal-2021-detecting,ramamoorthy2022memotion,sharma-etal-2020-semeval}. A semi-supervised learning approach using multimodal data containing both images and text could help ease the annotation process.

Therefore, our team realizes that the current urgent problem is to solve the issue of "memes sentiment analysis" to detect harmful content in the direction of research of "multimodal semi-supervised learning." By creating a multimodal semi-supervised learning approach, we can outline our contributions as follows:

\begin{itemize}
  \item Develop a pre-trained model, named Cross Modality Auto Encoder (CROM-AE), based on Contrastive Language-Image Pre-Training (CLIP) features that can be trained on small datasets without labels.
  \item Create a custom supervised model called RAW-N-COOK that incorporates the extracted features of both CROM-AE and CLIP, hence, utilizing knowledge from unlabeled data for a supervised model.
\end{itemize}

\section{Related Works}
Tasks involving multimodal content, consisting of both images and text, typically rely on the success of visual-linguistic pre-train models such as VisualBERT~\cite{li2019visualbert} and CLIP~\cite{radford2021learning}. Especially CLIP, based on self-supervised learning, hence, is trained on a large-scale dataset including 300M pairs of image-text from the internet. Thus, it may have learned popular concepts of visual and linguistic features on the internet, including memes. By utilizing CLIP encoders and feature interaction between modalities, in 2022, Kumar and Nanadakumar proposed the Hate-CLIPper architecture~\cite{karthik2022hate}. This model achieved the highest performance on the Hateful Memes Challenge dataset, obtaining an ROC-AUC score of 85.80, which surpassed human performance of 82.62 on the same dataset. Previously, the top solutions for Hateful Memes Challenge were based on VisualBERT or its variants~\cite{kiela2021hateful}. In other meme sentiment analysis competitions~\cite{fersini-etal-2022-semeval,ramamoorthy2022memotion}, successful approaches have typically employed supervised learning based on either VisualBERT or CLIP~\cite{fersini-etal-2022-semeval,patwa2022findings}.

Various studies on semi-supervised learning in multimodal data, such as images, text, and other modalities, have been conducted for different objectives. For example, Hu et al.~\cite{Hu_Zhu_Peng_Lin_2020} utilized feature projection on an embedding space and implemented cross-modal retrieval tasks to retrieve either text by image or vice versa. Sunkara et al.~\cite{Sunkara2020} utilized a large unlabelled audio and text corpus to pre-train modality encoders and fused the encoder's output to train punctuation prediction in conversational speech. Liang et al.~\cite{liang2020semi} worked on emotion recognition in videos by extracting visual, acoustic, and lexical signals from both labeled and unlabeled videos. Their end-to-end model simultaneously performs an auto-encoder task on the entire data and an emotion classification task on the latent representations of the labeled data's modalities. In the domain of memes, Gunti et al.~\cite{Gunti_Ramamoorthy_Patwa_Das_2022} tried to embed images and words in the same space by training a Siamese network that receives a pair of image-word belonging to a meme. As a result, they make the image embedding of a meme have a semantic meaning driven by word embedding. This way, they demonstrate how valuable unlabeled meme data can be used. Although these studies have different tasks and modalities compared to ours, they have inspired us with possible methods for applying semi-supervised learning on multimodal data.

In relation to our interest in implementing semi-supervised classification for images and text, there was a state-of-the-art (SOTA) study by Yang et al.~\cite{ijcai2019p568} in 2019. They proposed a method called Comprehensive Semi-Supervised Multimodal Learning (CMML), which utilized unlabeled data to strike a balance between consistency and diversity among modalities through the introduction of diversity and consistency measurements. Diversity measurements were optimized to increase the diversity among modalities' predictions, while consistency measurements were optimized to minimize disagreement among them. CMML achieved competitive results on large-scale multimodal datasets like FLICKR25K~\cite{10.1145/1460096.1460104}, IAPR TC-12~\cite{ESCALANTE2010419}, NUS-WIDE~\cite{10.1145/1646396.1646452}, and MS-COCO~\cite{10.1007/978-3-319-10602-1_48}. However, it is a difficult method to optimize because the loss function includes multiple supervised losses and regularized unsupervised losses. Additionally, it depends on the vector of multi-label predictions, making it only available for multi-label classification tasks.

\section{Methodology}
\subsection{Overview}
\begin{figure}
\includegraphics[width=\textwidth]{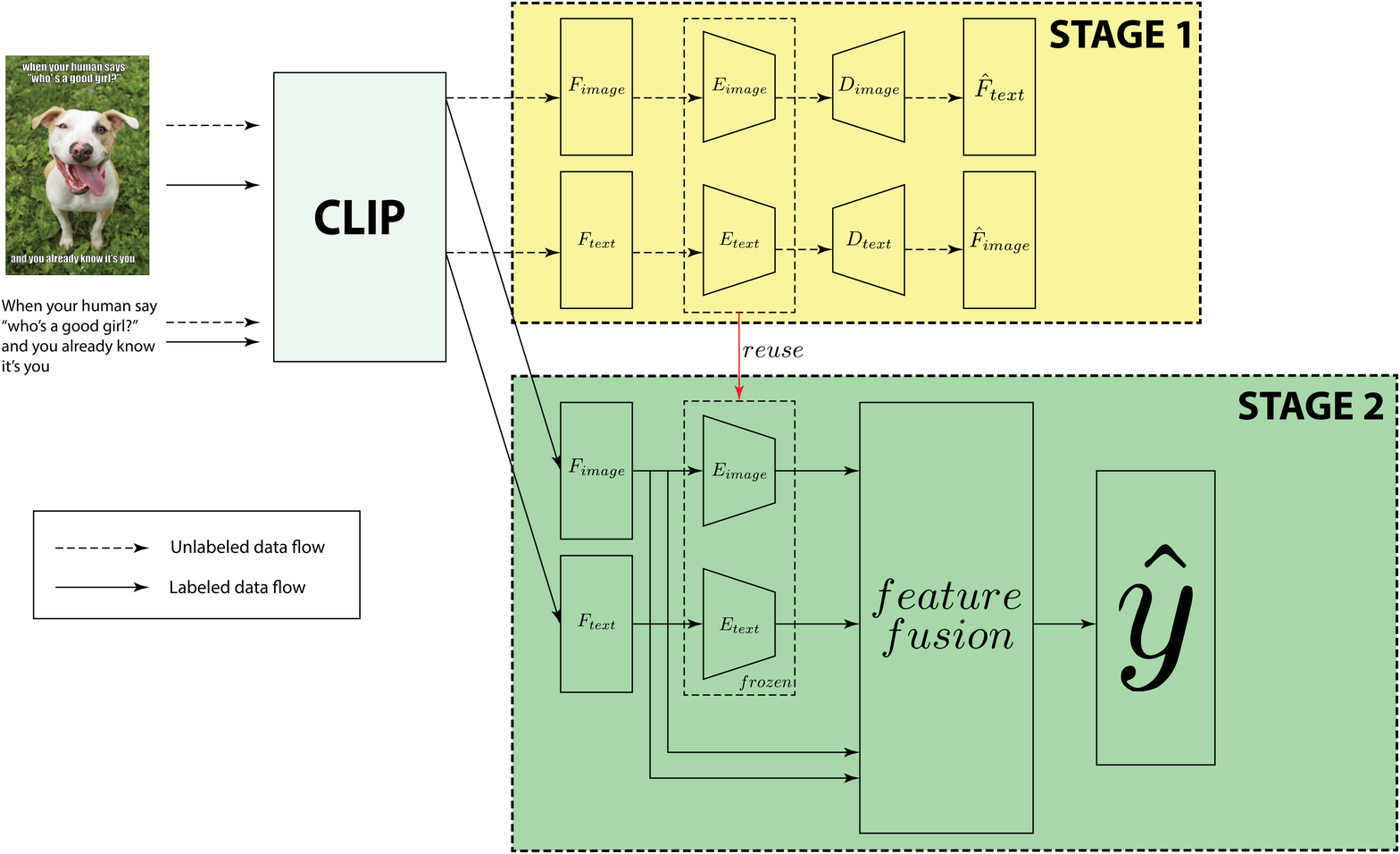}
\caption{
The overall architecture of our semi-supervised learning approach
}
\label{fig1}
\end{figure} 
In Fig.~\ref{fig1}, the initial step involves inputting a pair of (image, text) into the CLIP model's image and text encoders to extract CLIP feature vectors $(F_{image},F_{text})$. The process then proceeds in two stages:

\textbf{Stage 1} - Unsupervised Pre-training: An unsupervised pre-training phase where the Cross Modality Auto Encoder (CROM-AE) is trained. CROM-AE uses one modality's CLIP feature to predict the other modality's CLIP feature, i.e., image feature $F_{image}$ predicts text feature $F_{text}$ and vice versa. This pre-training phase is carried out using only unlabeled data.

\textbf{Stage 2} - Supervised Fine-tuning: A new model is designed for learning the classification task on labeled data. First, the pre-trained encoders of CROM-AE are frozen to extract new representations from the original CLIP features. Then, both the new representations (cooked features) and the original CLIP features (raw features) are fused to predict the classification target. The resulting model is called the Raw and Cooked Features Classification Model (RAW-N-COOK).

We call our 2-stages method SemiMemes. The following subsections, ~\ref{CROMAE} and ~\ref{RAWNCOOK}, will discuss the model used in each stage in detail.



\subsection{Cross Modality Auto Encoder (CROM-AE) - Stage 1}\label{CROMAE}
We define the Cross Modality Auto Encoder (CROM-AE) as a model that uses one modality to reconstruct the other modality. Specifically, there are two CROM-AE models, $AE_{image}$ and $AE_{text}$, where $AE_{image}$ takes CLIP features $F_{image}$ as input and predicts $\hat{F}_{text}$ as output, and $AE_{text}$ takes $F_{text}$ as input and predicts $\hat{F}_{image}$ as output. Formally, we can represent this as follows:

\begin{equation}
\hat{F}_{text} = AE_{image}(F_{image})
\end{equation}\label{eq:CROMAE_IMAGE}
\begin{equation}
\hat{F}_{image} = AE_{text}(F_{text})
\end{equation}\label{eq:CROMAE_TEXT}

Here, $F_{image}$ and $F_{text}$ are the CLIP features of the image and text, respectively, and $\hat{F}_{image}$ and $\hat{F}_{text}$ are the estimations of $AE_{text}$ and $AE_{image}$, respectively.

\begin{figure}
\includegraphics[width=\textwidth]{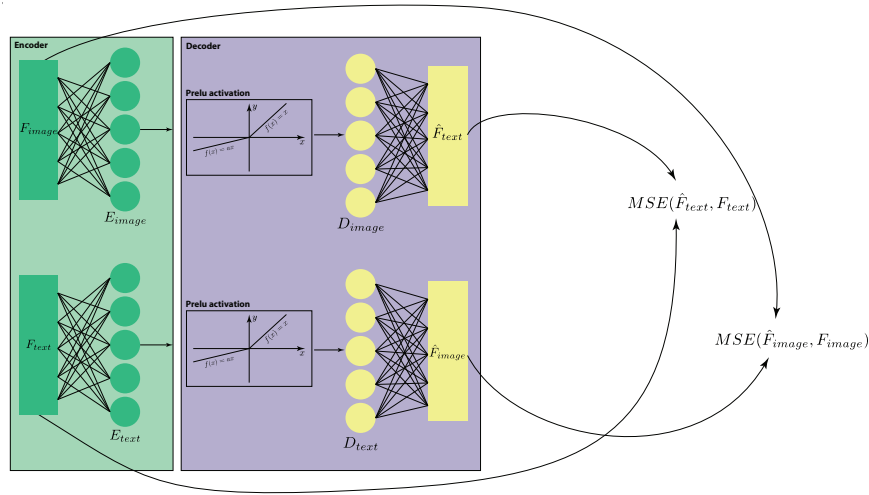}
\caption{
The pipeline of two CROM-AEs
}
\label{fig2}
\end{figure}

In practice, both auto-encoders have the same underlying architecture consisting of $Linear > PReLU > Linear$ where all the linear layers have the same dimensions of 768. We use PReLU~\cite{he2015delving} instead of the popular ReLU activation function to ensure that the model learns the negative values of the encoder's output. This is helpful in later stage. The encoder linear layer and decoder linear layer of each modality are denoted as $E_{k}$ and $D_{k}$ with $k \in {image, text}$ in Fig.~\ref{fig2}.

CROM-AE can be used to capture the underlying distribution of each modality for semi-supervised learning. The latent representations of images are driven by the distribution of the remaining text $p(text)$, which may contain the information of the posterior distribution $p(y|text)$, where $y$ is the supervised classification target and vice versa~\cite{ouali2020overview}. We exclude all labeled data when training the CROM-AE models to prevent introducing new bias and variance to the labeled training, validation, and test sets. The two CROM-AE models are trained separately using the Mean-Square-Error loss function.

\begin{equation}
    \mathcal{L}_{AE_{image}}=MSE(\hat{F} _{text},F_{text})
\end{equation}\label{eq:Loss_AE_Image}
\begin{equation}
    \mathcal{L}_{AE_{text}}=MSE(\hat{F} _{image},F_{image})
\end{equation}\label{eq:Loss_AE_Text}

\subsection{Raw and Cooked Features Classification Model (RAW-N-COOK) - Stage 2}\label{RAWNCOOK}

\begin{figure}
\includegraphics[width=\textwidth]{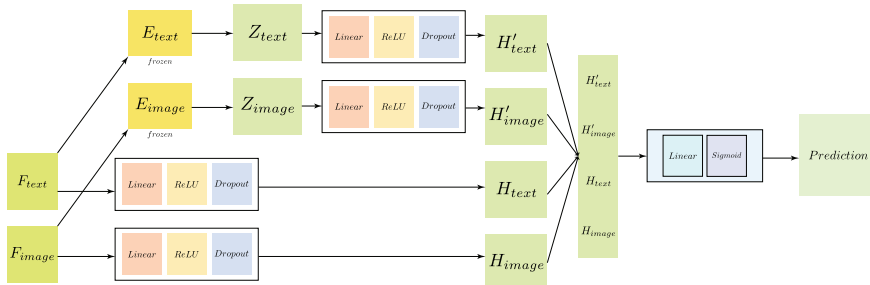}
\caption{
The architecture of RAW-N-COOK model for classification
}
\label{fig3}
\end{figure}

RAW-N-COOK is a classification model that incorporates both learned latent representation from CROM-AE and the original CLIP features as follows. Firstly, we take only the encoder part $E_{image}$, $E_{text}$ of two pre-trained CROM-AE models and freeze them. Then, both CLIP features $F_{image}$ and $F_{text}$ go through their corresponding CROM-AE encoder $E_{image}$,$E_{text}$ to obtain latent extracted features $Z_{image}$, $Z_{text}$. Then, four vectors: $F_{image}$,$F_{text}$, $Z_{image}$, $Z_{text}$ are projected to four 256-length vectors by a simple sequence of layers: $Linear > ReLU > Dropout$, then concatenate to obtain a 1024-length vector. The concatenated vector goes through the last $Linear$ layer to learn the classification target. The flow is described in Fig.~\ref{fig3}.

Our intuition is that $Z_{image}$, and $Z_{text}$ are informative features because they were learned on a large unlabeled dataset. Therefore, encoders are frozen to keep what CROM-AE learned on unlabeled data. However, both $Z_{image}$ and $Z_{text}$ were driven by different tasks, if we use only $Z_{image}$ and $Z_{text}$ for classification, it is not too powerful. Therefore, we decided to fuse $Z_{image}$ and $Z_{text}$ (cooked features) with the original features outputted from CLIP $F_{image}$,$F_{text}$ (raw features).

Recall the PReLU discussion in ~\ref{CROMAE}, if we use ReLU in the decoder, negative values on $Z_{image}$, and $Z_{text}$ are not learned by CROM-AE that will become noise in the classification model, which makes the model harder to learn on the classification task. Therefore, in CROM-AE, we choose PReLU.

\section{Experimental Results}
\subsection{Datasets}
\subsubsection{Multimedia Automated Misogyny Identification (MAMI)} is SemEval-2022 task 5's dataset~\cite{fersini-etal-2022-semeval}. It is designed to identify inappropriate memes towards women using both images and texts. The task consists of 2 sub-tasks: sub-task A and sub-task B. Sub-task A is a binary classification problem that aims to classify memes with hateful content into 2 categories: misogyny and not misogyny. Sub-task B involves identifying memes with misogynistic content, which incorporates identifying more fine-grained classes: stereotyping, shame, protest, and violence. This sub-task has 4 binary labels and is formulated as a multi-label classification problem. The MAMI dataset contains 10K memes for training and 1K memes for testing. We further select 2K samples from the training set to use as a validation set, the remaining samples of the training set are divided into labeled training set and unlabeled training set  In this study, we focus on sub-task B.
\subsubsection{Hateful Memes} is in Hateful Memes Challenge~\cite{NEURIPS2020_1b84c4ce}, constructed by Facebook, which focuses on detecting hate speech in multimodal memes. It is a binary classification task in which the labels of each meme need to be identified as either hateful or not hateful. The entire "seen" set consists of exactly 10K memes, which have been divided into 5\%, 10\%, and 85\% for the development set, test set, and training set, respectively. Additionally, the dataset includes "unseen" sets hidden during most of the challenge evaluation. We do not consider the "unseen" sets in this study.
\subsection{Setup}
\subsubsection{Preprocessing} As being in done~\cite{hakimov-etal-2022-tib} for the MAMI dataset, we undertook text preprocessing for SemiMems and CMML-CLIP through the following sequence of actions: elimination of text with mixed-in URLs, removal of non-ASCII characters, conversion of all characters to lowercase, and deletion of punctuation. In addition, we resized the image to be a square image with dimensions of 224x224 to correspond with the pre-train model's size. As to the Hateful Memes dataset, we just resized images to fit the input size of the CLIP encoder.

\subsubsection{Supervised Loss}
Due to the imbalance of MAMI datasets, which exists a large number of negative numbers compared to positive ones, we use the focal version of distribution balanced loss (DB loss)~\cite{wu2020distribution} to optimize the multi-label classification problem as described in the sub-task B of the dataset. The focal DB loss can be written as follows: 
\begin{equation}
    \mathcal{L}_{DB-focal}(x^{k},y^{k})=-\frac{1}{C}\sum^{C}_{i=0}\hat{r}^k_i[(1-p_+)^{\gamma}y^k_ilog(p_+)+\frac{1}{\lambda}p^\gamma_-(1-y^k_i)log(1-p_-)] 
\end{equation}
\begin{equation}
    p_- = \frac{1}{1+e^{-\lambda(z^k_i-\nu_i)}}
\end{equation}
\begin{equation}
    p_+ = \frac{1}{1+e^{-(z^k_i-\nu_i)}}
\end{equation}
\begin{equation}
    \nu_i = \kappa\hat{b}_i
\end{equation}
\begin{equation}
    \hat{b}_i = -log(\frac{1}{p_i}-1)
\end{equation}
Where C is the number of classes, $\hat{r}^k_i$ is the re-balanced weight as defined in ~\cite{wu2020distribution},
$p_-$, $p_+$ is the negative probability and positive probability, respectively, $z^k_i$ is the logit of model, $\gamma$ is the focusing parameter,
$\lambda$ is the scale factor that regularizes the loss gradient of negative samples, $\nu_i$ is the class-specific bias, $\hat{b}_i$ is the estimated bias variable, $\kappa$ is the scale factor of $\hat{b}_i$, and $p_i$ is the class prior probability which is equal to $\frac{Number\,sample\,of\,class\,i}{total\,number\,of\,training\,sample}$. We set $\gamma=2$, $\lambda=5$, $\kappa=0.1$ in our implementation.

For the Hateful Memes dataset, we use the traditional BCE loss to optimize the binary classification task.  
\subsubsection{Experimental setting and hyperparameters} To experiment, we randomly create a labeled set and an unlabeled set for training from the original training set. We experiment on 3 scenarios, each of which has the proportion of labeled set equal to 5\%, 10\%, and 30\% of the original dataset. We use Adam optimizer with an initial learning rate of 1e-4 and 0.25 for the initial value of $a$ in PReLU activation. During training, we reduce the learning rate after each epoch by StepLR scheduler with different $\gamma_{scheduler}$
for each scenario as shown in Table~\ref{tab1}. The number of pre-training epochs, fine-tuning epochs and batch size are 200, 200 and 40, respectively. For the Stage 1 on MAMI, we use L2 regularization with 1e-4 weight decay. The model is trained on NVIDIA-RTX 6000.
\subsubsection{Baselines} We compare SemiMemes with the following baselines, including SOTA methods of multimodal semi-supervised learning and multimodal supervised learning. 

\begin{table}
\centering
\caption{Values of $\gamma_{scheduler}$ for each setting.}\label{tab1}
\begin{tabular}{|l|l|l|}
\hline
 Dataset & Labeled ratio &  $\gamma_{scheduler}$ \\
\hline
\multirow{3}{*}{MAMI} & 5\% & 0.93  \\
& 10\% & 0.9 \\ 
& 30\% & 0.85 \\
\hline
\multirow{3}{*}{Hateful Memes} & 5\% & 0.96  \\
& 10\% & 0.96 \\ 
& 30\% & 0.96
\\
\hline
\end{tabular}
\end{table}

\begin{enumerate}
    \item Multimodal Semi-supervised
    \begin{itemize}
        \item \textbf{CMML-CLIP}: We find that using only the original CMML model, the performance is significantly low on the MAMI dataset, so we replace CMML's encoder with CLIP encoder for more robust representations.
    \end{itemize}
    \item Multimodal supervised learning 
    \begin{itemize}
        \item \textbf{TIB-VA}~\cite{hakimov-etal-2022-tib}: is the top winner solution in the MAMI competition 2022. The model uses CLIP encoders as backbones and the training follows multitasking objectives for both sub-task A and sub-task B. TIB-VA achieved 1st place on sub-task B.
        \item \textbf{Hate-CLIPper}~\cite{https://doi.org/10.48550/arxiv.2210.05916}: is the current SOTA model on the Hateful Memes dataset. There are 2 types of fusions that Hate-CLIPper proposed, which are align-fusion and cross-fusion. We run the model with cross-fusion and 1 pre-output layer as this configuration achieves the best result on Hateful Memes.
    \end{itemize}
We use the ViT-Large-Patch14 version for the CLIP encoder. We train all other models also in 200 epochs, and use the last checkpoint to evaluate the score on the test set. For supervised models, we just use labeled data in each scenario. For MAMI, we use the weighted F1 score as it is the standard metric used in the competition, for Hateful Memes' evaluation, the area under the receiver operating characteristic curve (AUROC) metric is used. 
    
\end{enumerate}

\subsection{Results}
\subsubsection{MAMI benchamrk}
As can be seen in Table~\ref{tab2}, SemiMemes's validation and test scores are all higher than the other 2 models on 3 label settings. In the 5\% label setting, SemiMemes achieves 0.693 on the test set which is more than that of CMML-CLIP and TIB-VA, by 0.034 and 0.039, respectively. These performance gaps decrease as the labeled samples increase, to 0.024 for SemiMemes versus CMMLP-CLIP and 0.023 for SemiMemes versus TIB-VA, in the 10\% labeled setting. In the 30\% labeled setting, the performance gap of SemiMemes with respect to CMMLP-CLIP  is 0.017, while the performance gap of SemiMemes with respect to TIB-VA increases to 0.031. This may demonstrate that, with a large number of unlabeled examples, our method can be more advantageous than other SOTA methods.
\begin{table}
\centering
\caption{Weighted-average F1-Measure on Validation and Test Set of MAMI dataset.}\label{tab2}
    \begin{tabular}{lcccccc}
    \toprule
    \multirow{2}{*}{Models} &
    \multicolumn{2}{c}{5 (\%)} &
    \multicolumn{2}{c}{10 (\%)} &
    \multicolumn{2}{c}{30 (\%)} \\
    & {Val} & {Test} & {Val} & {Test} & {Val} & {Test} \\
    \midrule
SemiMemes & \textbf{0.693} & \textbf{0.6782} & \textbf{0.7258}   & \textbf{0.7113} & \textbf{0.7520} & \textbf{0.7413} \\
CMML-CLIP & 0.6778 & 0.6438 & 0.717 & 0.6878 & 0.7313 & 0.7242  \\
TIB-VA & 0.68  & 0.6392 & 0.6992 & 0.6886  & 0.7095 & 0.7104  \\
\bottomrule
\end{tabular}
\end{table}

\subsubsection{Hateful Memes benchmark}
As shown in Table~\ref{tab3}, SemiMemes is also capable of achieving higher scores than Hate-CLIPper in all settings. The results of SemiMemes are not very higher than Hate-CLIPper on both development and test sets. Specifically, for the test set, the performance gaps are 0.004, 0.009, and 0.003 for 5\%, 10\%, and 30\% settings, respectively. However, when taking into account the number of parameters in each method, the total parameters of our models, CROM-AE and RAW-N-COOK included, are significantly lower than that of Hate-CLIPper, with about 3.1 million parameters compared to 1.5 billion parameters of Hate-CLIPper. Therefore, SemiMemes can save significant training resources, while still assuring high performance.
\begin{table}
\centering
\caption{AUROC on Dev Seen and Test Seen Set of Hateful Memes.}\label{tab3}
    \begin{tabular}{lccccccccc}
    \toprule
    \multirow{2}{*}{Models} &
    \multirow{2}{*}{Trainable params} &
    \multicolumn{2}{c}{5 (\%)} &
    \multicolumn{2}{c}{10 (\%)} &
    \multicolumn{2}{c}{30 (\%)} \\
    & &{Dev} & {Test} & {Dev} & {Test} & {Dev} & {Test} &
     \\
    \midrule

SemiMemes & \textbf{3.1M} & \textbf{0.6897} & \textbf{0.7011} & \textbf{0.7061}   & \textbf{0.7281} & \textbf{0.7399} & \textbf{0.7765} \\

Hate-CLIPper (cross) & 1.5B & 0.6652  & 0.6973  & 0.6827 & 0.7196 & 0.7030  & 0.7731 \\

\bottomrule
\end{tabular}
\end{table}

\section{Conclusion}
In conclusion, this research highlights the prevalence of internet memes and the need to sentiment analyze their underlying meanings for censoring harmful content. To achieve this, the proposed multimodal semi-supervised learning approach, SemiMemes, utilizes a combination of CROM-AE and RAW-N-COOK models to leverage the vast amount of unlabeled memes on the internet and mitigate the difficulties of the annotation process. This approach surpasses other state-of-the-art models on two datasets, demonstrating its effectiveness in identifying misogynistic and hateful memes in low-labeled data settings.

\bibliographystyle{splncs04}
\bibliography{refs}

\begin{thebibliography}{10}
\providecommand{\url}[1]{\texttt{#1}}
\providecommand{\urlprefix}{URL }
\providecommand{\doi}[1]{https://doi.org/#1}

\bibitem{borzsei2013makes}
B{\"o}rzsei, L.K.: Makes a meme instead. The Selected Works of Linda
  B{\"o}rzsei pp. 1--28 (2013)

\bibitem{10.1145/1646396.1646452}
Chua, T.S., Tang, J., Hong, R., Li, H., Luo, Z., Zheng, Y.: Nus-wide: A
  real-world web image database from national university of singapore. In:
  Proceedings of the ACM International Conference on Image and Video Retrieval.
  CIVR '09, Association for Computing Machinery, New York, NY, USA (2009).
  \doi{10.1145/1646396.1646452}

\bibitem{ESCALANTE2010419}
Escalante, H.J., Hernández, C.A., Gonzalez, J.A., López-López, A., Montes,
  M., Morales, E.F., {Enrique Sucar}, L., Villaseñor, L., Grubinger, M.: The
  segmented and annotated iapr tc-12 benchmark. Computer Vision and Image
  Understanding  \textbf{114}(4),  419--428 (2010).
  \doi{10.1016/j.cviu.2009.03.008}, special issue on Image and Video Retrieval
  Evaluation

\bibitem{fersini-etal-2022-semeval}
Fersini, E., Gasparini, F., Rizzi, G., Saibene, A., Chulvi, B., Rosso, P.,
  Lees, A., Sorensen, J.: {S}em{E}val-2022 task 5: Multimedia automatic
  misogyny identification. In: Proceedings of the 16th International Workshop
  on Semantic Evaluation (SemEval-2022). pp. 533--549. Association for
  Computational Linguistics, Seattle, United States (Jul 2022).
  \doi{10.18653/v1/2022.semeval-1.74}

\bibitem{pmlr-v5-goldberg09a}
Goldberg, A., Zhu, X., Singh, A., Xu, Z., Nowak, R.: Multi-manifold
  semi-supervised learning. In: van Dyk, D., Welling, M. (eds.) Proceedings of
  the Twelth International Conference on Artificial Intelligence and
  Statistics. Proceedings of Machine Learning Research, vol.~5, pp. 169--176.
  PMLR, Hilton Clearwater Beach Resort, Clearwater Beach, Florida USA (16--18
  Apr 2009)

\bibitem{Gunti_Ramamoorthy_Patwa_Das_2022}
Gunti, N., Ramamoorthy, S., Patwa, P., Das, A.: Memotion analysis through the
  lens of joint embedding (student abstract). Proceedings of the AAAI
  Conference on Artificial Intelligence  \textbf{36}(11),  12959--12960 (Jun
  2022). \doi{10.1609/aaai.v36i11.21616}

\bibitem{hakimov-etal-2022-tib}
Hakimov, S., Cheema, G.S., Ewerth, R.: {TIB}-{VA} at {S}em{E}val-2022 task 5: A
  multimodal architecture for the detection and classification of misogynous
  memes. In: Proceedings of the 16th International Workshop on Semantic
  Evaluation (SemEval-2022). pp. 756--760. Association for Computational
  Linguistics, Seattle, United States (Jul 2022).
  \doi{10.18653/v1/2022.semeval-1.105}

\bibitem{he2015delving}
He, K., Zhang, X., Ren, S., Sun, J.: Delving deep into rectifiers: Surpassing
  human-level performance on imagenet classification. In: Proceedings of the
  IEEE international conference on computer vision. pp. 1026--1034 (2015)

\bibitem{Hu_Zhu_Peng_Lin_2020}
Hu, P., Zhu, H., Peng, X., Lin, J.: Semi-supervised multi-modal learning with
  balanced spectral decomposition. Proceedings of the AAAI Conference on
  Artificial Intelligence  \textbf{34}(01),  99--106 (Apr 2020).
  \doi{10.1609/aaai.v34i01.5339}

\bibitem{10.1145/1460096.1460104}
Huiskes, M.J., Lew, M.S.: The mir flickr retrieval evaluation. In: Proceedings
  of the 1st ACM International Conference on Multimedia Information Retrieval.
  p. 39–43. MIR '08, Association for Computing Machinery, New York, NY, USA
  (2008). \doi{10.1145/1460096.1460104}

\bibitem{karthik2022hate}
Karthik~Kumar, G., Nandakumar, K.: Hate-clipper: Multimodal hateful meme
  classification based on cross-modal interaction of clip features. arXiv
  e-prints pp. arXiv--2210 (2022)

\bibitem{kiela2021hateful}
Kiela, D., Firooz, H., Mohan, A., Goswami, V., Singh, A., Fitzpatrick, C.A.,
  Bull, P., Lipstein, G., Nelli, T., Zhu, R., et~al.: The hateful memes
  challenge: Competition report. In: NeurIPS 2020 Competition and Demonstration
  Track. pp. 344--360. PMLR (2021)

\bibitem{NEURIPS2020_1b84c4ce}
Kiela, D., Firooz, H., Mohan, A., Goswami, V., Singh, A., Ringshia, P.,
  Testuggine, D.: The hateful memes challenge: Detecting hate speech in
  multimodal memes. In: Larochelle, H., Ranzato, M., Hadsell, R., Balcan, M.,
  Lin, H. (eds.) Advances in Neural Information Processing Systems. vol.~33,
  pp. 2611--2624. Curran Associates, Inc. (2020)

\bibitem{https://doi.org/10.48550/arxiv.2210.05916}
Kumar, G.K., Nandakumar, K.: Hate-clipper: Multimodal hateful meme
  classification based on cross-modal interaction of clip features (2022).
  \doi{10.48550/ARXIV.2210.05916}

\bibitem{li2019visualbert}
Li, L.H., Yatskar, M., Yin, D., Hsieh, C.J., Chang, K.W.: Visualbert: A simple
  and performant baseline for vision and language. arXiv preprint
  arXiv:1908.03557  (2019)

\bibitem{liang2020semi}
Liang, J., Li, R., Jin, Q.: Semi-supervised multi-modal emotion recognition
  with cross-modal distribution matching. In: Proceedings of the 28th ACM
  International Conference on Multimedia. pp. 2852--2861 (2020)

\bibitem{10.1007/978-3-319-10602-1_48}
Lin, T.Y., Maire, M., Belongie, S., Hays, J., Perona, P., Ramanan, D.,
  Doll{\'a}r, P., Zitnick, C.L.: Microsoft coco: Common objects in context. In:
  Fleet, D., Pajdla, T., Schiele, B., Tuytelaars, T. (eds.) Computer Vision --
  ECCV 2014. pp. 740--755. Springer International Publishing, Cham (2014).
  \doi{10.1007/978-3-319-10602-1{\_}48}

\bibitem{ouali2020overview}
Ouali, Y., Hudelot, C., Tami, M.: An overview of deep semi-supervised learning.
  arXiv preprint arXiv:2006.05278  (2020)

\bibitem{patwa2022findings}
Patwa, P., Ramamoorthy, S., Gunti, N., Mishra, S., Suryavardan, S., Reganti,
  A., Das, A., Chakraborty, T., Sheth, A., Ekbal, A., et~al.: Findings of
  memotion 2: Sentiment and emotion analysis of memes. In: Proceedings of
  De-Factify: Workshop on Multimodal Fact Checking and Hate Speech Detection,
  CEUR (2022)

\bibitem{pramanick-etal-2021-detecting}
Pramanick, S., Dimitrov, D., Mukherjee, R., Sharma, S., Akhtar, M.S., Nakov,
  P., Chakraborty, T.: Detecting harmful memes and their targets. In: Findings
  of the Association for Computational Linguistics: ACL-IJCNLP 2021. pp.
  2783--2796. Association for Computational Linguistics, Online (Aug 2021).
  \doi{10.18653/v1/2021.findings-acl.246}

\bibitem{radford2021learning}
Radford, A., Kim, J.W., Hallacy, C., Ramesh, A., Goh, G., Agarwal, S., Sastry,
  G., Askell, A., Mishkin, P., Clark, J., et~al.: Learning transferable visual
  models from natural language supervision. In: International conference on
  machine learning. pp. 8748--8763. PMLR (2021)

\bibitem{ramamoorthy2022memotion}
Ramamoorthy, S., Gunti, N., Mishra, S., Suryavardan, S., Reganti, A., Patwa,
  P., DaS, A., Chakraborty, T., Sheth, A., Ekbal, A., et~al.: Memotion 2:
  Dataset on sentiment and emotion analysis of memes. In: Proceedings of
  De-Factify: Workshop on Multimodal Fact Checking and Hate Speech Detection,
  CEUR (2022)

\bibitem{sharma-etal-2020-semeval}
Sharma, C., Bhageria, D., Scott, W., PYKL, S., Das, A., Chakraborty, T.,
  Pulabaigari, V., Gamb{\"a}ck, B.: {S}em{E}val-2020 task 8: Memotion analysis-
  the visuo-lingual metaphor! In: Proceedings of the Fourteenth Workshop on
  Semantic Evaluation. pp. 759--773. International Committee for Computational
  Linguistics, Barcelona (online) (Dec 2020).
  \doi{10.18653/v1/2020.semeval-1.99}

\bibitem{Sunkara2020}
Sunkara, M., Ronanki, S., Bekal, D., Bodapati, S., Kirchhoff, K.: Multimodal
  semi-supervised learning framework for punctuation prediction in
  conversational speech. In: Interspeech 2020 (2020)

\bibitem{wu2020distribution}
Wu, T., Huang, Q., Liu, Z., Wang, Y., Lin, D.: Distribution-balanced loss for
  multi-label classification in long-tailed datasets. In: Computer Vision--ECCV
  2020: 16th European Conference, Glasgow, UK, August 23--28, 2020,
  Proceedings, Part IV 16. pp. 162--178. Springer (2020).
  \doi{10.1007/978-3-030-58548-8{\_}10}

\bibitem{ijcai2019p568}
Yang, Y., Wang, K.T., Zhan, D.C., Xiong, H., Jiang, Y.: Comprehensive
  semi-supervised multi-modal learning. In: Proceedings of the Twenty-Eighth
  International Joint Conference on Artificial Intelligence, {IJCAI-19}. pp.
  4092--4098. International Joint Conferences on Artificial Intelligence
  Organization (7 2019). \doi{10.24963/ijcai.2019/568}

\end{thebibliography}
\end{document}